\documentclass[conference]{IEEEtran}
\IEEEoverridecommandlockouts
\usepackage{cite}
\usepackage{amsmath,amssymb,amsfonts}
\usepackage{algorithmic}
\usepackage{graphicx}
\usepackage{textcomp}
\usepackage{xcolor}
\usepackage{multirow}
\usepackage{float}
\def\BibTeX{{\rm B\kern-.05em{\sc i\kern-.025em b}\kern-.08em
    T\kern-.1667em\lower.7ex\hbox{E}\kern-.125emX}}
\begin{document}


\title{Incremental Label Distribution Learning with Scalable Graph Convolutional Networks}
\author{\IEEEauthorblockN{
   Ziqi Jia\IEEEauthorrefmark{2}\IEEEauthorrefmark{3}\textsuperscript{a}\thanks{\textsuperscript{a}equal contribution},
    Xiaoyang Qu\IEEEauthorrefmark{2}\textsuperscript{a},
    Chenghao Liu\IEEEauthorrefmark{3},
    and Jianzong Wang\thanks{*Corresponding author: Jianzong Wang, jzwang@188.com}\IEEEauthorrefmark{2}\IEEEauthorrefmark{1}
    \IEEEauthorblockA{\IEEEauthorrefmark{2}Ping An Technology (Shenzhen) Co., Ltd., Shenzhen, China}
    \IEEEauthorblockA{\IEEEauthorrefmark{3}Tsinghua Shenzhen International Graduate School, Tsinghua University, Shenzhen, China}
}
}

\maketitle

\begin{abstract}
Label Distribution Learning (LDL) is an effective approach for handling label ambiguity, as it can analyze all labels at once and indicate the extent to which each label describes a given sample. Most existing LDL methods consider the number of labels to be static. However, in various LDL-specific contexts (e.g., disease diagnosis), the label count grows over time (such as the discovery of new diseases), a factor that existing methods overlook. Learning samples with new labels directly means learning all labels at once, thus wasting more time on the old labels and even risking overfitting the old labels. At the same time, learning new labels by the LDL model means reconstructing the inter-label relationships. How to make use of constructed relationships is also a crucial challenge. To tackle these challenges, we introduce Incremental Label Distribution Learning (ILDL), analyze its key issues regarding training samples and inter-label relationships, and propose Scalable Graph Label Distribution Learning (SGLDL) as a practical framework for implementing ILDL. Specifically, in SGLDL, we develop a New-label-aware Gradient Compensation Loss to speed up the learning of new labels and represent inter-label relationships as a graph to reduce the time required to reconstruct inter-label relationships. Experimental results on the classical LDL dataset show the clear advantages of unique algorithms and illustrate the importance of a dedicated design for the ILDL problem.
\end{abstract}

\section{Introduction}
Label distribution learning (LDL) \cite{geng2016label} a powerful framework for dealing with complex prediction tasks where uncertainty or ambiguity exists about the correct label or set of labels. The output of LDL is the probability distribution of labels for each input instance, indicating the degree to which each label is associated with the input instance. This output form differs from both Single Label Learning (SLL)\cite{wang2020deep} and Multi-Label Learning (MLL)\cite{mahapatra2022multi}. LDL has recently achieved remarkable successes in various domains, such as sentiment analysis, recommendation systems, fraud detection, and disease diagnosis\cite{jia2019label}.

Generally, most existing LDL methods are modelled with a static scenario where labels are fixed. However, the real world is often dynamic, and the model needs to learn new labels to keep performing well. To handle such a setting, existing LDL methods typically abandon the old model and create a new one. In other words, no related work currently exists to enable the LDL model to learn new labels while retaining its knowledge of old labels. Based on the characteristics of LDL, we believe that the main reason LDL cannot be adapted to dynamic scenarios is the \textbf{label attention trap}\cite{hu2018few}. 

\begin{figure}[t]
    \centering
    \includegraphics[width=0.47\textwidth]{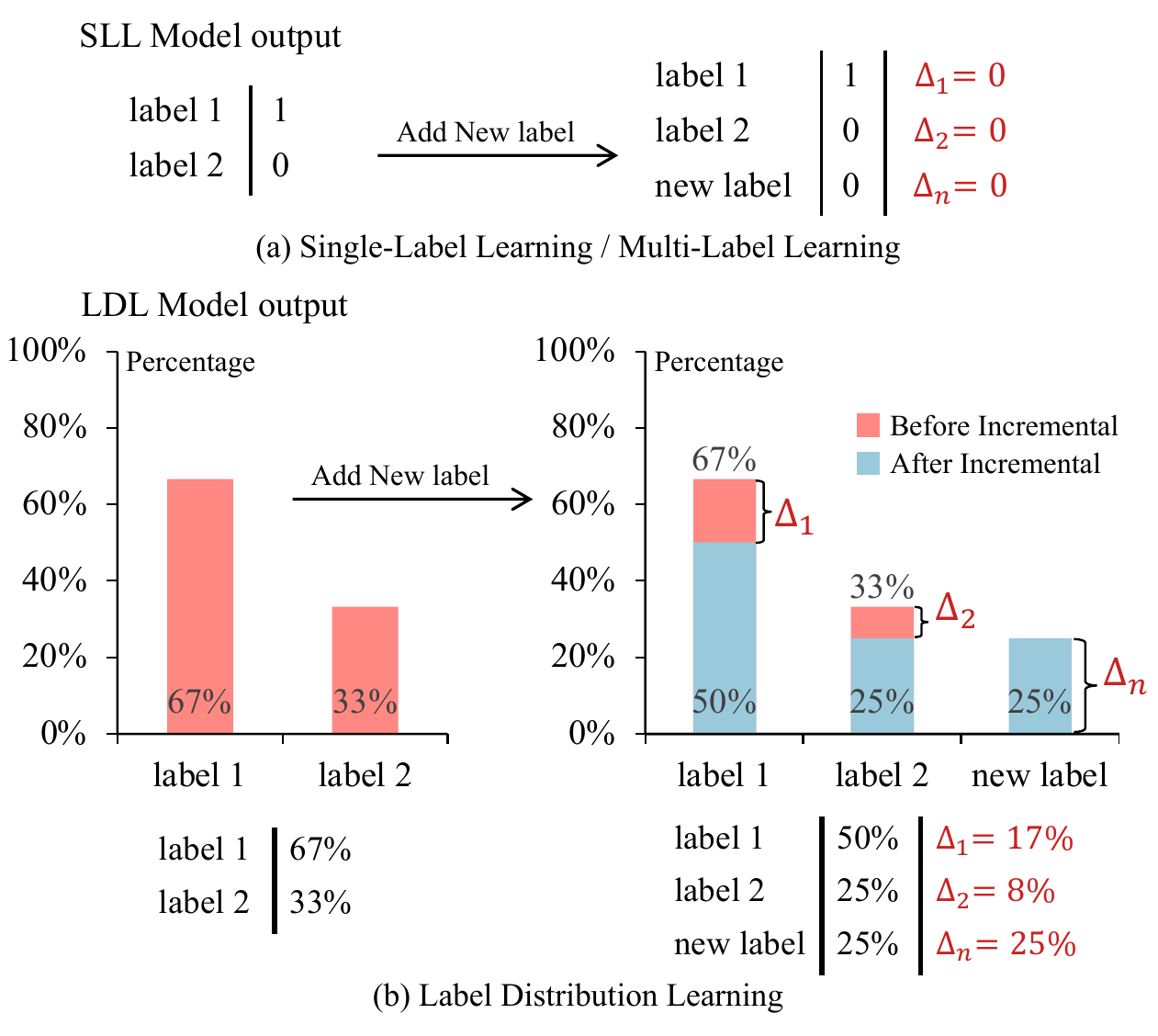}
    \caption{A toy example to show how the Label Distribution Learning model changes the prediction value of the old labels after learning a new label. As we can see, SLL / MLL and LDL judge the same label distribution very differently, predisposing their corresponding models to react very differently when learning new labels.}
\end{figure}

\begin{figure}[t]
    \centering
    \includegraphics[width=0.47\textwidth]{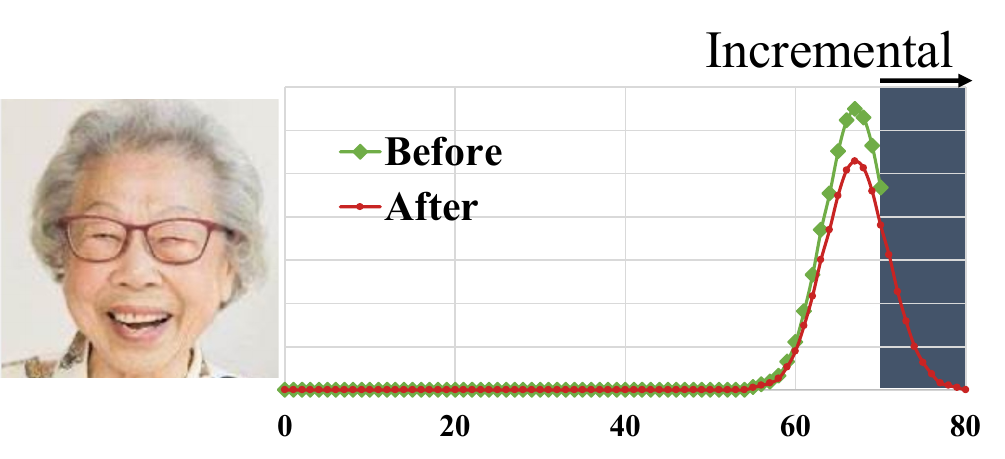}
    \caption{An example of predicting age distribution. The model outputs different distributions for the same image before and after incremental learning in the model's label space.}
\end{figure}

Label attention trap refers to the fact that the LDL model learns all the labels simultaneously rather than focusing more on the new labels. As shown in Fig. 1, When new labels are added to the label space, the prediction value of old labels scales down. Then the LDL model will give nearly equal attention to all labels, thus re-establishing inter-label relationships to ensure accurate model output. In fact, \textbf{the inter-label relationship of old labels is not changed}, so all the model needs to do is find a proper way to add new labels into the relationship and pay more attention to new labels.

To address the label attention trap, we address a challenging LDL problem named Incremental Label Distribution Learning (ILDL)\cite{liu2023learnable}. In the ILDL setting, the model can learn new labels at any time, and the new sample used for training has all old and new labels. ILDL requires the model to study continuously, with speed and precision. To better comprehend the ILDL problem, we use the medical diagnosis as a possible example.\cite{cermelli2022incremental} In medical diagnosis, doctors use LDL to assist in determining a patient's condition, and the output of the LDL model is the probability that the patient has certain diseases. The emergence of new diseases can cause the output of an already trained model to be inaccurate or even incorrect because the model has not learned about the new disease.\cite{wang2024label} In this case, existing LDL methods will likely suffer from the label attention trap under the emergence of new diseases.

A direct way to tackle studying new labels continuously in the ILDL setting is to simply integrate LDL and Class-incremental learning (CIL)\cite{zhou2024class} together. However, this strategy is likely to let the model fall into the label attention trap, which should be prevented. Additionally, the LDL model produces a real-valued vector instead of a binary one, as it assigns multiple continuous description scores to each label. Without considering the difference between MLL and LDL and the inter-relationships among old and new labels, this simple integration strategy could further exacerbate the label attention trap due to the error establishment of new label inter-relationships.

To solve these challenges in LDL, we introduce Scalable Graph Label Distribution Learning (SGLDL), which effectively addresses Label attention trap using a scalable label graph (SLG)\cite{jin2024gldl} and a new-label-aware gradient compensation loss. Specifically, to make inter-label relationships intuitive and easy-changeable, we use SLG to represent them and separate them from the feature extraction part of the model. SLG uses an expandable correlation matrix to record the relationships. During learning the new labels, SLG will add the relations about the new labels into the matrix while ensuring that the old relations remain unchanged, thus solving the problem that the model constructs incorrect inter-label relations based on incorrect attention allocation. However, more than improving the algorithm from this one aspect is needed, so we propose new-label-aware gradient compensation loss to facilitate learning new labels by balancing the learning speed of old labels with that of new labels. By comparing it with direct learning and some methods that combine CIL with LDL, SGLDL performs better on the dataset we set under the CIL task. \textit{The major contributions of this paper are summarized as follows: }
\begin{itemize}
    \item We tackle a real-world FL problem, Incremental Label Distribution Learning (ILDL), where the primary challenge lies in overcoming the label attention trap caused by incorrect attention allocation and maintaining the integrity of inter-label relationships.
    \item To address the ILDL problem, We develop a novel Scalable Graph Label Distribution Learning (SGLDL) algorithm. which using a scalable label graph (SLG) and a new-label-aware gradient compensation loss to solve label attention trap.
    \item We offer a new idea of designing the label relations separately from the feature extraction in LDL. 
\end{itemize}

\section{Preliminary}
\subsection{Problem Definition}
In the standard framework of label distribution learning, the following notations are commonly used: an instance variable is represented by \( x \), with the specific \( i \)-th instance noted as \( x_i \). A label variable is denoted by \( y \), and \( y_j \) refers to the specific value of the \( j \)-th label. The degree to which \( y \) describes \( x \) is expressed as \( d_x^y \). For an instance \( x_i \), its label distribution is given by \( D_i = \{d^{y_1}_{x_i}, d^{y_2}_{x_i}, \ldots, d^{y_c}_{x_i}\} \), where \( c \) is the count of all possible labels.


We then extend conventional LDL to Incremental Label Distribution Learning . ILDL focus on continual learning for a sequence of Label Distribution Learning  tasks $\mathcal{T}=\{\mathcal{T}^t\}^T_{t=1}$, where $T$ denotes the task number. The $t$-th task $\mathcal{T}^t=\{x^t_i,D^t_i\}_{i=1}^{N^t}$ consists of $N^t$ pairs of instances $x^t_i$ and their label distribution $D_i^t=\{d^{y_1}_{x_i^t},d^{y_2}_{x_i^t},\cdots,d^{y_{c^t}}_{x_i^t}\}$, where $c^t$ is the number of possible labels in $t$th task. the label space of $t$-th task is $\mathcal{Y}^t$, including new classes $\mathcal{Y}^t_\text{new}$ that are different from $\mathcal{Y}^{t-1}$ in previous $t-1$ tasks. It should be note that in ILDL settings, $D^t_i$ always including all labels that model has learned.

The target of ILDL is to learn a conditional probability mass function $f(x;\theta)$ from $\mathcal{T}$, $\theta$ is a set of parameters containing two parts: $\theta_G$ and $\theta_C$.  After learning all tasks in the sequence, we seek to minimize the generalization error on all tasks:
\begin{equation}
    R\left(f(x;\theta)\right)=\sum_{t=1}^T \mathbb{E}_{\left(x^t, D^t\right) \sim \mathcal{T}^t} \mathcal{L}\left(f(x^t;\theta), D^t\right)
\end{equation}

\begin{figure}[t]
    \centering
    \includegraphics[width=0.45\textwidth]{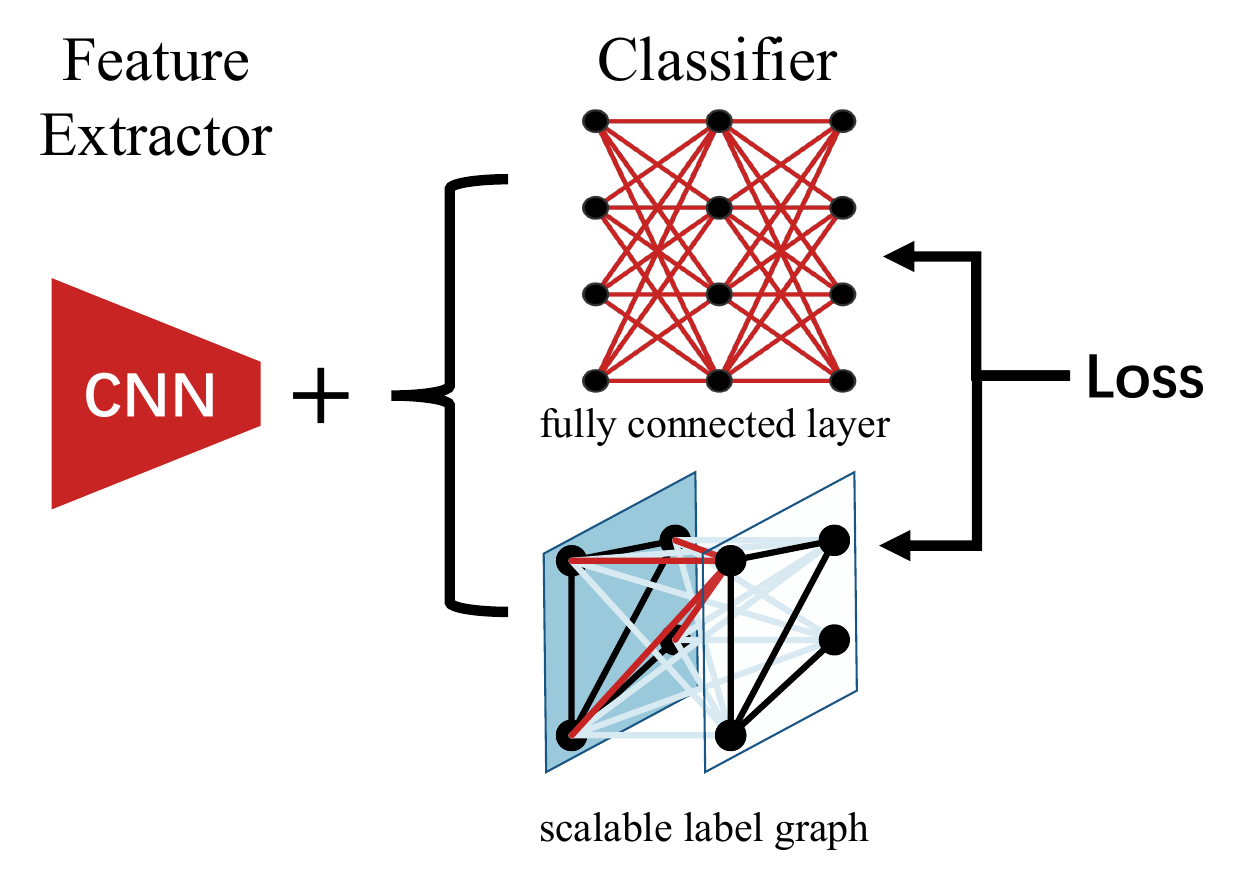}
    \caption{The variation of parameter modification caused by different classifiers in solving the ILDL problem. The deep red region indicates the parameters that need to be modified. Traditional fully connected layer need to modify all parameters due to the output characteristics of LDL. But for SGLDL, only the parameter values corresponding to the new label need to be modified, as only the knowledge in this part changed.}
\end{figure}

\subsection{Label Attention Trap}
Label attention trap in ILDL setting means that during LDL model learning new labels $\mathcal{Y}^t_\text{new}$ in $T^t$, the output $D^i$ is longer than $D^{i-1}$ since the label added. Unlike traditional class-incremental learning, as illustrated in Fig. 2, in ILDL, adding new labels leads to bias in the model's predictions for all labels. The classical LDL model consists of two parts: the feature extraction part and the classifier. The classifier is a fully connected layer of two or more layers. Due to the nature of the LDL algorithm, the classifier of the LDL model considers more inter-label interactions in its output, which means that the main information in the classifier part is the inter-label relationships. For ILDL, learning new labels leads to a bias in the model's prediction of almost all labels. Ordinary LDL models use methods such as back-propagation to correct for the bias, which modifies almost all the parameters(Fig. 3) . In contrast, in the ILDL setting, since the old inter-label relationships are unchanged, the parameters in this part do not need to be modified, and the focus should be shifted to constructing relationships between the new labels and the old labels and between the new labels and the new labels. We refer to this approach, where the model incorrectly iterates over all parameters rather than only those related to the new label, as the label attention trap.

\begin{figure}[t]
    \centering
    \includegraphics[width=0.45\textwidth]{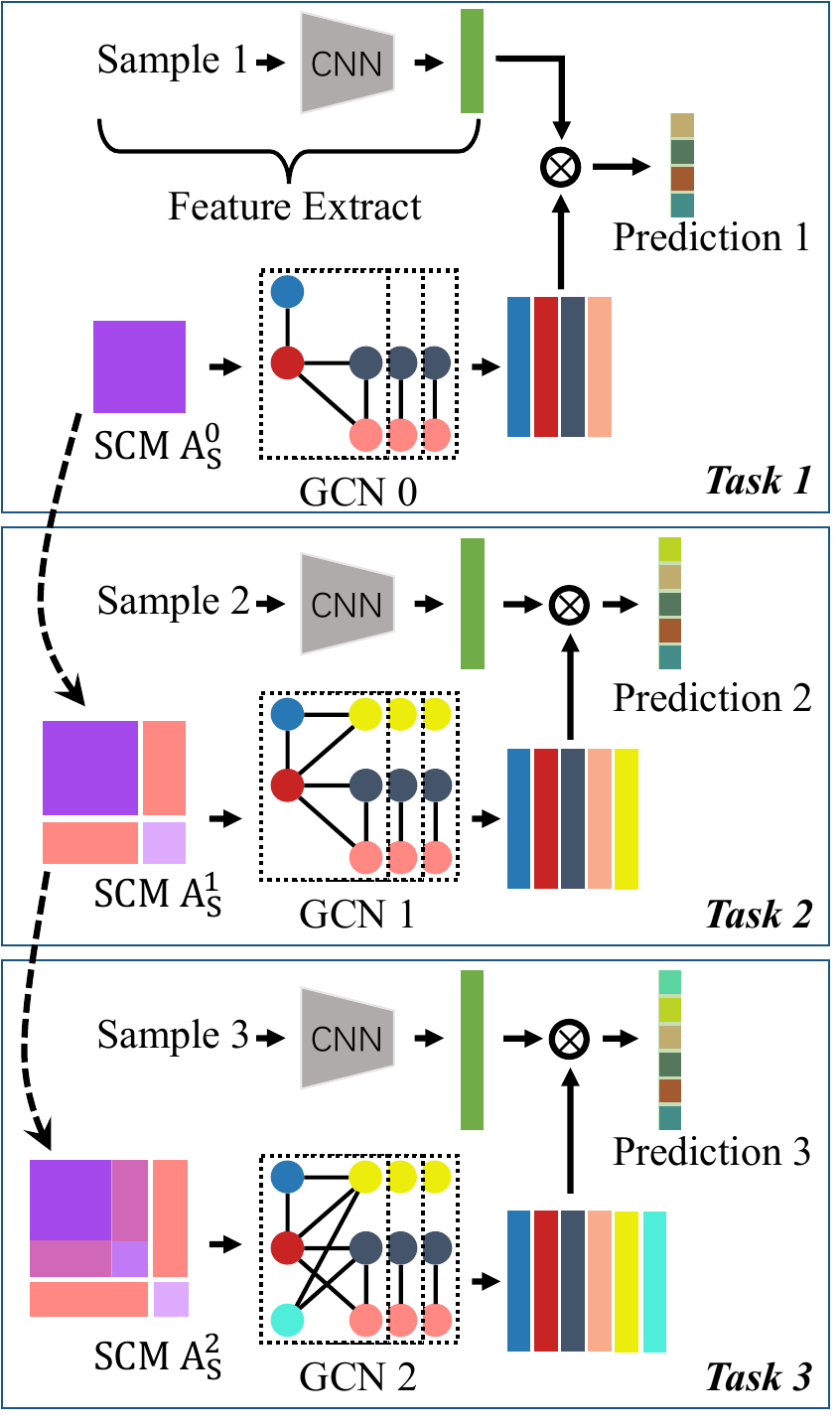}
    \caption{The framework of SGLDL involves feeding samples into a feature extractor (CNN) to extract features while generating the corresponding GCN through SCM to ensure effective incremental learning. The SCM is saved, and each incremental learning step extends the classifier (GCN) based on the previous SCM.}
\end{figure}

\section{Method}
To address the ILDL requirements, Graph-based Incremental Label Distribution Learning (SGLDL) tackles the label attention trap via a scalable label graph (SLG) and a new-label-aware gradient compensation loss. The SGLDL framework, illustrated in Fig. 4, consists of two primary modules, \textit{i.e.}, the feature extraction module and the graph convolutional network (GCN) classifier.

\subsection{Scalable Correlation Matrix}
To address the issue of label ambiguity, the LDL algorithm focuses on inter-label relationships in the design process and is reflected in the classifier. Specifically, LDL starts with designing the loss function, using a specific loss function to ensure that the model learns the inter-label relationship adequately. For ILDL, adding new labels can bias the model's predictions for all labels. Using fully connected layers as the classifier will cause the model to relearn the old labels while learning the new ones, which will slow the learning of the new labels and cause the model to overfit the old ones.

To solve the mentioned problem, we propose the scalable label graph (SLG), which uses a GCN to record inter-label relationships. GCN works based on the correlation matrix $\mathbf{A}$ to propagate information between nodes. Therefore, how to construct $\mathbf{A}$ is the key problem of GCN. In most applications, the correlation matrix is pre-defined. However, there is no related work in the field of LDL. In this paper, we use a data-driven approach to build the correlation matrix. That is, we define the correlation between labels by mining the co-occurrence patterns of the labels in the dataset.

The key module of SLG is scalable correlation matrix (SCM)\cite{alqallaf2002scalable}, which provides the inter-label relationships among all $\mathcal{Y}^t$. Specifically, We construct SCM $\mathbf{A}_S^t$ for $t$-th ($t>1$) task in an online fashion denoted as
\begin{equation}
    \mathbf{A}_S^t=\left[\begin{array}{cc}
\mathbf{A}_S^{t-1} & \mathbf{E}^t \\
\mathbf{R}^t & \mathbf{M}^t
\end{array}\right]=\left[\begin{array}{cc}
\text{old}\leftrightarrow\text{old} & \text {old}\to\text{new} \\
\text{new}\to\text{old} & \text{new} \leftrightarrow \text{new}
\end{array}\right]
\end{equation}
in which we set four submatrices $\mathbf{A}_S^{t-1}$ and $\mathbf{M}^t$, $\mathbf{E}^t$ and $\mathbf{R}^t$ to represent inter-label relationships between old-and-old labels, new-and-new labels, old-to-new labels as well as new-to-old labels respectively. For the first task ($t=1$), $\mathbf{A}_S^1=\mathbf{M}^1$. For $t>1$, $\mathbf{A}_S^t\in\mathbb{R}^{|\mathcal{Y}^{t-1}|\times|\mathcal{Y}^{t-1}|}$. Since $\mathbf{A}_S^{t-1}$ can be obtained directly from the old task, we will focus on how to compute the other three submatrices in the SCM.

\textbf{New-and-New Submatrix }($\mathbf{M}^t\in\mathbb{R}^{|\mathcal{Y}^{t}_\text{new}|\times|\mathcal{Y}^{t}_\text{new}|}$) computes the inter-label relationships among new labels. Multi-label learning (MLL) use conditional probability $P(\mathcal{Y}_i|\mathcal{Y}_j)$ to model the relationships since the output of the MLL model is $(0/1)$, which is easy to count. But LDL outputs the corresponding percentages for all labels, you cannot count the output. So there is a need to establish another method to quantify the impact of one label on another. For an LDL output $D^{y_j}$, the specific two labels is also influenced by other labels. To escape this influence, for an instant $\{x_k,D_k\}$, we set
\begin{equation}\label{eq_mij}
\begin{array}{cc}
{m_{ij}}_k=\frac{d^{y_i}_k}{d^{y_j}_k} & (d^{y_j}_k\neq 0)
\end{array}
\end{equation}
And we calculate $\mathbf{M}^t$ using:
\begin{equation}\label{eq_N}
    \mathbf{M}^t_{ij}=\frac{1}{\bar{m}_{ij}}\sqrt{\frac{\sum_{k=1}^{{n^t}_{ij}}({m_{ij}}_k-\bar{m}_{ij})^2}{{n^t}_{ij}}}
\end{equation}
where ${n^t}_{ij}$ is the number of samples in which $d^{y_j}_k\neq 0$, $\bar{m}_{ij}$ is the average of ${m}_{ij}$.

The function of Eq.\ref{eq_mij} is to remove the influence of the other labels by calculating $d^{y_i}_k/d^{y_j}_k({~}={m_{ij}}_k)$ and $d^{y_j}_k/d^{y_j}_k(=1)$. So that we can compare the statistical properties of $m_{ij}$ with $1$, allowing us to study the effect of $d^{y_j}_k$ on $d^{y_j}_k$ directly. 

In Eq.\ref{eq_N} we compute the coefficient of variation of $m_{ij}$ to quantify the effect of $d^{y_j}_k$ on $d^{y_j}_k$ since the coefficient of variation can show the volatility of $m_{ij}$. The important reason that we just calculate the coefficient of variation of $m_{ij}$ is because $d^{y_j}_k/d^{y_j}_k$ always equal to $1$, and it is meaningless to compare with a distribution full of $1$. With Eq.\ref{eq_N}, we can know the inter-label relationship between $d^{y_j}_k$ and $d^{y_j}_k$.

\textbf{Old-to-New Submatrix }($\mathbf{E}^t\in\mathbb{R}^{|\mathcal{Y}^{t-1}|\times|\mathcal{Y}^{t}_\text{n ew}|}$) computes the influence on new labels created by old labels. For an instance $\{x^t_k,D^t_k\}$, the old model $f(x;\theta_{t-1})$ can output the label distribution of old labels $\mathring{D}^{t}_k=\{\mathring{d}^{y_1}_{x^t_k},\mathring{d}^{y_2}_{x^t_k},\cdots,\mathring{d}^{y_{c^{t-1}}}_{x^t_k}\}$. If we only use the information in task $\mathcal{T}^t$, then we cannot deal with the fact that $\mathcal{T}^t$ may be biased against the true information recorded in the old labels. So we choose to use the information in $\mathcal{T}^t$ and the old model to ensure that the knowledge we learn is as authentic as possible. Specifically, based on Eq.\ref{eq_mij}, we have

\begin{equation}\label{eq_eij}
\begin{array}{cc}
    {e_{ij}}_k=\frac{d^{y_i}_k}{\sqrt{(1-d^{y_i}_k)\times\mathring{d}^{y_j}_k\times d^{y_j}_k}} & (\mathring{d}^{y_j}_k\neq 0; d^{y_j}_k\neq 0;d^{y_i}_k\neq 1)
\end{array}
\end{equation}
In Eq.\ref{eq_eij}, $(1-d^{y_i}_k)$ is used to reduce $\mathring{d}^{y_j}_k$ for the old model cannot calculate the percentage of the new labels, which take a part in the true distribution. After solving ${e_{ij}}_k$, we have
\begin{equation}
    \mathbf{E}^t_{ij}=\frac{1}{\bar{e}_{ij}}\sqrt{\frac{\sum_{k=1}^{{n^t}_{ij}}({e_{ij}}_k-\bar{e}_{ij})^2}{{n^t}_{ij}}}
\end{equation}

\textbf{New-to-Old Submatrix }($\mathbf{R}^t\in\mathbb{R}^{|\mathcal{Y}^{t}_\text{new}|\times|\mathcal{Y}^{t-1}|}$) computes the influence from new labels to old labels. After the old model output $\mathring{D}^{t}_k$, we can get
\begin{equation}
\begin{array}{cc}
    {r_{ij}}_k=\frac{\sqrt{(1-d^{y_i}_k)\times\mathring{d}^{y_j}_k\times d^{y_j}_k}}{d^{y_i}_k} & (d^{y_i}_k\neq 0)
\end{array}
\end{equation}
and then we can calculate $\mathbf{R}^t$ by
\begin{equation}
    \mathbf{R}^t_{ij}=\frac{1}{\bar{r}_{ij}}\sqrt{\frac{\sum_{k=1}^{{n^t}_{ij}}({r_{ij}}_k-\bar{r}_{ij})^2}{{n^t}_{ij}}}
\end{equation}
After calculate $\mathbf{M}^t,\mathbf{E}^t$ and $\mathbf{R}^t$, we can get SCM $\mathbf{A}_S^t$ for task $\mathcal{T}^t$.

\subsection{Scalable Label Graph}
SCM record inter-label relationships among all labels in $\mathcal{Y}^t$. With SCM, we can leverage GCN to assist the prediction of convolutional neural network (CNN). Specially, we propose Scalabel Graph Convolutional Network (SGCN) to manage SCM. SGCN is learned to map this label graph into a set of inter-dependent object classifiers, and further help the prediction of CNN as shown in Eq.\ref{eq_pred}. SGCN is a two-layer stacked graph model. Based on the SCM $\mathbf{A}_S^t$, SGCN can capture class-incremental dependencies flexibly. Let the graph node be initialized by the Glove embedding[cite] namely $\mathbf{H}^{t,0}\in \mathbb{R}^{|\mathcal{Y}^t|\times d} $ where $d$ represents the embedding dimensionality. The graph presentation $\mathbf{H}^t$ in task $t$ is mapped by: 
\begin{equation}
\mathbf{H}^t=\operatorname{SGCN}\left(\mathbf{A}_S^t, \mathbf{H}^{t, 0}\right)
\end{equation}
As shown in Fig. 4, together with an CNN feature extractor, the label distribution for an image x will be predicted by
\begin{equation}\label{eq_pred}
    \hat{D}= \mathcal{S}\left(\operatorname{SGCN}\left(\mathbf{A}_S^t, \mathbf{H}^{t, 0}\right) \otimes \mathrm{CNN}(x)\right)
\end{equation}
where $\mathbf{A}_S^t$ denotes the SCM and $\mathbf{H}^{t, 0}$ is the initialized graph node. $\mathcal{S}(\cdot)$ is softmax function, for $\mathbf{a}=\{a_1,a_2,\cdots,a_n\}$,
\begin{equation}
    \mathcal{S}(\mathbf{a}) = \left\{\frac{\exp{a_1}}{\sum^n_{i=1}\exp{a_1}},\cdots,\frac{\exp{a_n}}{\sum^n_{i=1}\exp{a_n}}\right\}
\end{equation}
\subsection{New-label-aware Gradient Compensation Loss}
In SGLDL, we want to train a suitable model $f(x^t;\theta)$ for all labels. For ease of illustration, we have divided $\theta$ into two parts, $\theta_C$ for CNN feature extractor and $\theta_G$ for GCN classifier. To solve the label attention trap, we here propose a new-label-aware gradient compensation loss $\ell_{\mathrm{NC}}$ to accelerate learning process via re-weighting gradient propagation. Specifically, for a single instance $\{x^t_k,D^t_k\}$, we obtain a gradient measurement $\mathcal{G}^t_k$ with respect to the $y^j$-th $(y^j\in\mathcal{Y}^t)$ neuron $\mathcal{N}^t_{{y^j_k}}$ of the last embedding layer in $\theta^t_C$:
\begin{equation}
    \mathcal{G}^t_{y^j_k} = \frac{\partial\mathcal{D}_{\mathrm{CD}}({\hat{y}^j_k,{y^j_k)}}}{\partial\mathcal{N}^t_{{y^j_k}}}
\end{equation}

where $\mathcal{D}_{\mathrm{CD}}(\cdot,\cdot)$ is Canberra metric[cite] and $\hat{y}^j_k$ is the predicted value of label $y^j_k$ with $x^t_k$ as input. With this measurement, we perform separate gradient normalization for old and new labels, and utilize it to compensate new label learning. Given a mini-batch $\{x^t_k,D^t_k\}_{k=1}^b$, we define
\begin{equation}
    \mathcal{G}_n=\frac{1}{b\times(c^t-c^{t-1})}\sum_{k=1}^b\sum^{c^t}_{j=c^{t-1}+1} \left|\mathcal{G}^t_{y^j_k}\right|
\end{equation}
as the gradient means for new labels, where using $c^t-c^{t-1}$ to make sure the index of new labels. Thus, the new-label-aware gradient compensation loss $\ell_{\mathrm{NC}}$ is formulated as follows:
\begin{equation}
    \ell_{\mathrm{NC}}=\frac{1}{b}\sum^b_{k=1}\sum^{c^t}_{j=1}\frac{\left|\mathcal{G}^t_{y^j_k}\right|}{\bar{\mathcal{G}}^t_{y^j_k}}\cdot \mathcal{D}_{\mathrm{CD}}({\hat{y}^j_k,{y^j_k)}}
\end{equation}
where $\bar{\mathcal{G}}^t_{y^j_k}=\mathbb{I}_{y^j_k \in \mathcal{Y}_\text{new}^t}\cdot \mathcal{G}_n+\mathbb{I}_{y^j_k \in \mathcal{Y}^{t-1}}\cdot \left|\mathcal{G}^t_{y^j_k}\right|$. and $\mathbb{I}(\cdot)$ is the indicator function that if the subscript condition is true, $\mathbb{I}(\text{True})=1$; otherwise, $\mathbb{I}(\text{False})=0$.

\subsection{Other Additional Loss Functions}
Inspired by the distillation-based lifelong learning method \cite{hou2018lifelong}, we save the old model ($\theta^{t-1}$) and construct a distillation loss $\ell_{\mathrm{DT}}$ to distill the knowledge of old labels between old model and new model. Specially,
\begin{equation}
    \ell_{\mathrm{DT}}=-\sum^b_{k=1}\sum_{j=1}^{c^{t-1}}\hat{y}^j_k\ln{y^j_k}
\end{equation}

In the process of incremental learning, the label graph includes inter-label relationships and is one of the core elements of SGLDL. To deal with label attention trap, we propose SCM $\mathbf{A}_S^t$, but the label graph is not only composed of SCM. Suppose the learned embedding after task $t$ is stored as $G^t=\mathrm{SGCN}(\mathbf{A}_S^t,\mathbf{H}^{t, 0})$, $t>1$. We propose a relation ship-preserving loss $\ell_{\mathrm{RP}}$ as a constraint to the inter-label relationships:
\begin{equation}
    \ell_{\mathrm{RP}}=\sum_{j=1}^{\left|\mathcal{Y}^{t-1}\right|}\left\|\mathbf{G}_j^{t-1}-\mathbf{H}_j^t\right\|^2
\end{equation}
By minimizing $\ell_{\mathrm{RP}}$ with the partial constraint of old node embedding, the variation in AGCN parameters is limited. As a result, the forgetting of established label relationships is mitigated as ILDL proceeded. The final loss of model training is defined as
\begin{equation}
    \ell = \lambda_1\ell_{\mathrm{NC}} + \lambda_2\ell_{\mathrm{DT}} +
    \lambda_3\ell_{\mathrm{RP}}
\end{equation}

\section{Experiments}
\subsection{Datasets}
\textbf{The IMDB-WIKI dataset}\cite{rothe2018deep} is the largest publicly available dataset of facial images with gender and age labels for training purposes. The facial images in this dataset were crawled from the IMDB and Wikipedia websites. 461,871 facial images are available on the IMDB website and 62,359 on the Wikipedia website. Given our limited computing power, we randomly selected 240,000 images to generate three small datasets with 80,000 face images each. To use Deep Age Distribution Learning, we assumed that the age distribution obeyed a discrete Gaussian distribution, specified as age with mean $\mu$ and standard deviation $\sigma = 3$.


\begin{table}[t]
\centering
\caption{Metrics for ILDL algorithms}
\begin{tabular}{|l|l|l|}
\hline
 {} & Name & Formula \\ \hline
\multicolumn{1}{|c|}{\multirow{1}{*}{\rotatebox[origin=c]{90}{Distance}}} & \begin{tabular}[c]{@{}l@{}}Euclidean\\ Distance\end{tabular}  $\downarrow$        & $Dis_1=\sqrt{\sum\limits^L\limits_{j=1}(P_j-Q_j)^2}$      \\ \cline{2-3} 
\multicolumn{1}{|c|}{}                          & \begin{tabular}[c]{@{}l@{}}Kullback-Leibler\\ Divergence\end{tabular}   $\downarrow$ & $Dis_2=\sum\limits^L\limits_{j=1}\ln\frac{P_j}{Q_j}$      \\ \hline
\multirow{1}{*}{\rotatebox[origin=c]{90}{Similarity}} & Intersection $\uparrow$ & $Sim_1 = \sum\limits^L\limits_{j=1}\min(P_j,Q_j) $      \\ \cline{2-3} 
& Fidelity $\uparrow$    & $Sim_2 = \sum\limits^L\limits_{j=1}\sqrt{P_jQ_j}$ \\ \hline
\end{tabular}

\vspace{-0.6cm}
\label{t_dis}
\end{table}

\begin{table*}[t]
    \centering
    \caption{Performance comparisons on The IMDB-WIKI dataset measured by Euclidean Distance}
    \scalebox{1.25}{
    \begin{tabular}{|l|cccccccccc|}
    \hline
         \textbf{Methods} & \textbf{10} & \textbf{20} & \textbf{30} & \textbf{40} & \textbf{50} & \textbf{60} & \textbf{70} & \textbf{80} & \textbf{90} & \textbf{100} \\\hline
         Adam-LDL-SCL & 0.107 & 0.162 & 0.256 & 0.375 & 0.514 & 0.594 & 0.734 & 0.760 & 0.849 & 0.906\\
         iCaRL + Adam-LDL-SCL & 0.098 & 0.155 & 0.195 & 0.254 & 0.346 & 0. 372 & 0.433 & 0.444 & 0.490 & 0.580\\
         BiC +  Adam-LDL-SCL & 0.096 & 0.127 & 0.154 & 0.197 & 0.279 & 0.342 & 0.402 & 0.437 & 0.488 & 0.552\\
         SS-IL +  Adam-LDL-SCL & 0.099 & 0.146 & 0.187 & 0.233 & 0.272 & 0.317 & 0.399 & 0.441 & 0.472 & 0.595\\
         \hline
         SGLDL- w/oLNC& 0.097 & 0.121 & 0.164 & 0.217 & 0.297 & 0. 381 & 0.493 & 0.537 & 0.485 & 0.621\\
         SGLDL- w/oLDT& 0.099 & 0.136 & 0.182 & 0.223 & .281 &0. 394 & 0.485 & 0.523 & 0.498 &.635\\
         SGLDL- w/oLRP &0.096 &0.120 &0.148& 0.199&0.284&0.320&0.399&0.446&0.474&0.553\\
         \textbf{SGLDL}& \textbf{0.094} & \textbf{0.117} & \textbf{0.142} & \textbf{0.191} & \textbf{0.267} & \textbf{0.311} & \textbf{0.393} & \textbf{0.434} & \textbf{0.468} & \textbf{0.551}\\
         \hline
    \end{tabular}
    }

    \label{tab:CVcifar}
\end{table*}


\subsection{Settings}
For a fair comparison, the SGD optimizer, whose learning rate is 2.0, is used to train all models. We also design understandable methods combining class incremental learning methods and the state-of-the-art LDL method.

For the IMDB-WIKI dataset, we use EfficientNet-B0 as the backbone of all methods, and the SOTA LDL method used here is DLDL. Moreover, for the Human Gene dataset, we use multi-layer perception as the backbone of all methods, and the SOTA LDL method used here is Adam-LDL-SCL.

According to the requirements of incremental learning, it is necessary to correct the label distribution of test samples before inputting them into the model. This means that the label space of the samples should only contain the labels that the model has already learned. Since the label space of the test set corresponds to the label space in the multi-task learning (MTL) scenario, which includes all labels, the label distribution of the test set is adjusted before comparing it with the model's output. The labels that the model has not yet learned are removed, and the remaining labels' distribution values are geometrically normalized. This ensures that the sum of these labels' distributions is the same as the label distribution of the model's output, which equals to 1. After normalize, we can guarantee the proper evaluation of the model.

\subsection{Results and Analysis}
We divided the dataset into 10 subsets according to the requirements of ILDL, serving as 10 incremental learning tasks. The experimental results are presented in the form of "Calculated results (rank)". Rank refers to the predictive performance of all ILDL algorithms on each metric, where lower values indicate better performance. Additionally, the best result is highlighted in bold. The experimental results are reported in Table \ref{tab:CVcifar}.
There are four ILDL algorithms in the experiment. Adam-LDL-SCL\cite{jia2019label} enhances label distribution learning by leveraging the label correlation of local samples and employing the adaptive learning rate optimization method Adam. iCaRL\cite{rebuffi2017icarl} learns new classes by leveraging exemplar samples that approximate class centers and knowledge distillation, enabling the recognition of new classes and updating model representations. BiC\cite{wu2019large} effectively mitigates the impact of data imbalance by employing a small validation set to calibrate the bias introduced by new classes in the classifier layer during incremental learning. SS-IL\cite{ahn2021ss} addresses the class score imbalance issue in incremental learning by employing a distinct softmax output layer combined with task-specific knowledge distillation.
\begin{figure}[H]
\centering
    \includegraphics[width=0.6\linewidth]{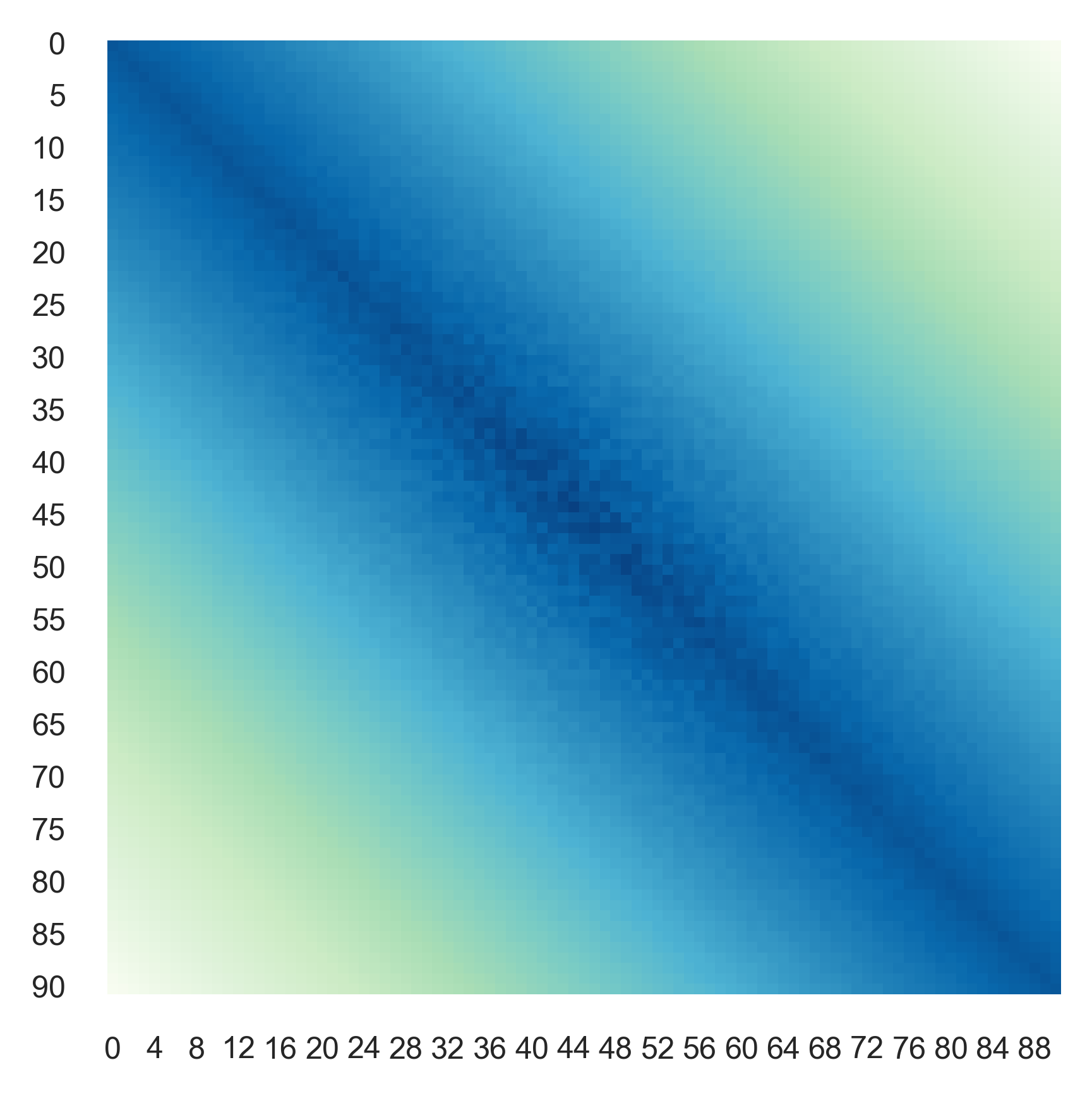}
    \caption{SCM visualization on IMDB-WIKI dataset}
    \label{hot1}
\end{figure}
By analyzing the results in Table \ref{tab:CVcifar}, We can observe that in most cases, We can observe that algorithms performing well in conventional CIL (such as iCaRL) still have some effectiveness in ILDL. However, it is evident that BiC outperforms iCaRL and SS-IL, primarily due to BiC's emphasis on compensating for class imbalance. At the same time, both iCaRL and SS-IL fall into the 'Label Attention Trap' as they prioritize new labels. Our proposed SGLDL consistently delivers top-tier performance across diverse scenarios. Fig. 5 provides visual evidence of the superior capabilities of the Scalable Label Graph in capturing and leveraging intricate inter-label relationships throughout the incremental learning process. This enhanced ability to model label dependencies contributes significantly to SGLDL's robust performance.

\subsection{Ablation Studies and Analysis}
As can be seen from Table \ref{tab:CVcifar}, we conducted ablation studies to investigate the role of each loss in our model. SGLDL-w/oLNC, SGLDL- w/oLDT, SGLDL- w/oLRP denote the performence of SGLDL without using $\ell_\text{NC}$, $\ell_\text{DT}$ and $\ell_\text{RP}$. where SGLDL- w/oLDT utilize $\ell_\text{DT}$ and the knowledge distillation proposed in iCaRL for a replacement. And SGLDL-w/oLRP use one layer Graph to make sure $\ell_\text{RP}$ is useless. Compared with SGLDL, the performance of SGLDL-w/oLNC, SGLDL- w/oLDT and SGLDL- w/oLRP degrades evidently. This validates the effectiveness of the collaborative work of the individual loss functions. It also demonstrates the existence of 'Label Attention Trap' in ILDL and the effectiveness of $\ell_\text{NC}$ in addressing this issue. 
\section{Related Works}

\subsection{Label Distribution Learning}
Label Distribution Learning assigns a distribution of labels to each instance, capturing relationships beyond single-label classification and enabling connections to multiple labels. LDL has been successfully applied in fields like age estimation\cite{gao2018age,wen2020adaptive} and emotion analysis\cite{jia2019facial}.

Custom algorithms have been developed specifically to align with the unique properties of LDL, and these approaches have been widely explored. IIS-LLD \cite{geng2013facial} uses a maximum entropy model to assign a label distribution to each face image, allowing each sample to contribute to learning both its actual age and nearby ages.
SCE-LDL \cite{10.5555/3060832.3060937} enhances the learning of complex image features by introducing an energy function with sparsity constraints, addressing the challenge of extracting sufficient useful information that traditional LDL methods often face. Additionally, LDL has also incorporated several innovative concepts. LDSVR \cite{10.5555/2832581.2832738} applies multi-output support vector machines to improve accuracy in predicting audience rating distributions, addressing LDL’s multivariable output challenges. DLDL\cite{gao2017deep} combines convolutional neural networks with label distribution learning, using Kullback-Leibler divergence to capture label ambiguity and improve generalization on small datasets by leveraging label correlations in both feature and classifier learning.

\subsection{Incremental Learning}
Incremental learning enables a model to continually acquire new tasks or categories while preserving previously learned knowledge. A major challenge here is catastrophic forgetting, where the model loses prior information as it learns new tasks\cite{wang2024incprompt,de2021continual}. To address this, several approaches have been developed. Learning without forgetting uses Knowledge Distillation Loss to maintain performance on prior tasks using only new task data, eliminating the need to revisit old data \cite{li2017learning}. iCaRL preserves knowledge by storing exemplar samples of earlier categories for reuse \cite{rebuffi2017icarl}. Piggyback mitigates forgetting by learning binary masks for each new task, selectively activating network weights without altering the original model \cite{mallya2018piggyback}.

\subsection{Graph Convolutional Network}
Graph Convolutional Networks (GCNs) are effective for learning representations from graph-structured data, capturing relationships and dependencies that are often missed in grid-like datasets\cite{wu2019simplifying}. By generalizing the concept of convolution to graphs, GCNs aggregate features from neighboring nodes, creating embeddings that reflect both individual node attributes and the overall graph structure. This is achieved through a message-passing mechanism, where nodes exchange information with neighbors, followed by neural network layers that transform the aggregated information, enabling GCNs to learn rich node-level, edge-level, or graph-level representations.

\section{Conclusion}
In this paper, we present a novel problem called Incremental Label Distribution Learning (ILDL) and analyze the most crucial issue in ILDL, the 'Label Attention Trap'. To tackle this issue, we present SLG and a New-label-aware Gradient Compensation Loss, which help reduce the negative effect of introducing new labels on the existing ones and counteract the 'Label Attention Trap'. Furthermore, we propose SGLDL, which combines SLG and New-label-aware Gradient Compensation Loss together. Comprehensive experiments on standard datasets validate the effectiveness of the proposed SGLDL algorithm.
\section{Acknowledgements}
This paper is supported by the Key Research and Development Program of Guangdong Province under grant No.2021B0101400003. Corresponding author is Jianzong Wang from Ping An Technology (Shenzhen) Co., Ltd.
\bibliographystyle{IEEEtran}
\bibliography{IEEEabrv}

\end{document}